\documentclass[conference]{IEEEtran}
\usepackage{cite}
\usepackage{amsmath,amssymb,amsfonts}
\usepackage{algorithmic}
\usepackage{graphicx}
\usepackage{textcomp}
\usepackage{xcolor}
\usepackage{booktabs}
\usepackage{url}
\usepackage{subcaption}
\usepackage[colorlinks=true, urlcolor=blue, linkcolor=blue]{hyperref}

\def\BibTeX{{\rm B\kern-.05em{\sc i\kern-.025em b}\kern-.08em
    T\kern-.1667em\lower.ex\hbox{E}\kern-.125emX}}

\begin{document}

\title{TBT-Former: Learning Temporal Boundary Distributions for Action Localization}

\author{\IEEEauthorblockN{Thisara Rathnayaka}
\IEEEauthorblockA{\textit{Dept. of Computer Science \& Engineering} \\
\textit{University of Moratuwa}\\
Sri Lanka \\
thisara.21@cse.mrt.ac.lk}
\and
\IEEEauthorblockN{Dr. Uthayasanker Thayasivam}
\IEEEauthorblockA{\textit{Dept. of Computer Science \& Engineering} \\
\textit{University of Moratuwa}\\
Sri Lanka \\
rtuthaya@cse.mrt.ac.lk}
}

\maketitle

\begin{abstract}
Temporal Action Localization (TAL) remains a fundamental challenge in video understanding, aiming to identify the start time, end time, and category of all action instances within untrimmed videos. While recent single-stage, anchor-free models like ActionFormer have set a high standard by leveraging Transformers for temporal reasoning, they often struggle with two persistent issues: the precise localization of actions with ambiguous or "fuzzy" temporal boundaries and the effective fusion of multi-scale contextual information. In this paper, we introduce the Temporal Boundary Transformer (TBT-Former), a new architecture that directly addresses these limitations. TBT-Former enhances the strong ActionFormer baseline with three core contributions: (1) a higher-capacity scaled Transformer backbone with an increased number of attention heads and an expanded Multi-Layer Perceptron (MLP) dimension for more powerful temporal feature extraction; (2) a cross-scale feature pyramid network (FPN) that integrates a top-down pathway with lateral connections, enabling richer fusion of high-level semantics and low-level temporal details; and (3) a novel boundary distribution regression head. Inspired by the principles of Generalized Focal Loss (GFL), this new head recasts the challenging task of boundary regression as a more flexible probability distribution learning problem, allowing the model to explicitly represent and reason about boundary uncertainty. Within the paradigm of Transformer-based architectures, TBT-Former advances the formidable benchmark set by its predecessors, establishing a new level of performance on the highly competitive THUMOS14 and EPIC-Kitchens 100 datasets, while remaining competitive on the large-scale ActivityNet-1.3. Our code is available \href{https://github.com/aaivu/In21-S7-CS4681-AML-Research-Projects/tree/main/projects/210536K-Multi-Modal-Learning_Video-Understanding}{here}.
\end{abstract}


\section{Introduction}
The ability to automatically understand human activities from video is a cornerstone of modern computer vision, with applications ranging from video summarization and surveillance to human-computer interaction \cite{b1}. A critical task in this domain is Temporal Action Localization (TAL), which requires not only classifying actions but also precisely identifying their temporal start and end points within long, untrimmed videos. The evolution of TAL methodologies has seen a significant shift from complex two-stage, proposal-based frameworks to more streamlined and efficient single-stage, anchor-free approaches \cite{b2, b3}.

A seminal work in this new paradigm is ActionFormer \cite{b2}, which demonstrated the remarkable efficacy of a minimalist, Transformer-based architecture. By treating every moment in a video as a potential action center and using local self-attention within a multi-scale encoder, ActionFormer achieved state-of-the-art results without relying on pre-defined anchors or explicit proposal generation stages \cite{b2}. Despite its success, the ActionFormer framework and similar models exhibit two primary limitations that motivate our work.

First, the problem of \textbf{imprecise boundary regression} persists. Standard regression heads, typically optimized with an L1 (Least Absolute Deviations) or DIoU (Distance-Intersection over Union) loss, predict a single, deterministic offset for an action's start and end times. This formulation is inherently inflexible and struggles when action boundaries are ambiguous—for instance, when an action begins or ends gradually. In such real-world scenarios, forcing the model to predict a single correct timestamp can lead to noisy training signals and inaccurate localization \cite{b4}.

Second, the mechanism for \textbf{multi-scale feature fusion} can be improved. The feature pyramid in ActionFormer is constructed in a purely feed-forward manner, where strided operations generate progressively coarser feature maps \cite{b2}. This architecture lacks an explicit mechanism for coarser, semantically rich features to inform and refine the finer-grained features at higher temporal resolutions. This limits the model's ability to integrate global context with precise local details, which is crucial for localizing actions of vastly different durations \cite{b2}.

To address these challenges, we propose the Temporal Boundary Transformer (TBT-Former), a new model that introduces three targeted architectural advancements upon the ActionFormer baseline.
\begin{enumerate}
    \item A \textbf{Scaled Transformer Backbone}: We increase the model's representational capacity by scaling up the Transformer encoder, employing 16 attention heads and expanding the MLP hidden dimension by a factor of 6x. This allows the model to capture more intricate and long-range temporal dependencies.
    \item A \textbf{Cross-Scale Feature Pyramid (CS-FPN)}: Inspired by refined Feature Pyramid Network (FPN) structures \cite{b5}, we incorporate a top-down pathway with lateral connections. This architecture facilitates a bidirectional flow of information, enriching fine-grained temporal features with high-level semantic context from coarser scales.
    \item A novel \textbf{Boundary Distribution Regression (BDR) Head}: This is our main contribution. Drawing inspiration from the core philosophy of Generalized Focal Loss (GFL) \cite{b6}, we reframe boundary prediction. Instead of regressing a single offset value, our BDR head learns a flexible probability distribution over a range of possible temporal locations. This allows the model to explicitly represent boundary uncertainty, leading to more robust and accurate localization, especially for actions with ambiguous transitions.
\end{enumerate}

Through extensive experimentation, TBT-Former demonstrates superior performance, setting a new state-of-the-art on the THUMOS14 \cite{b7} and EPIC-Kitchens 100 \cite{b8} datasets and achieving highly competitive results on ActivityNet-1.3 \cite{b9}. Our work shows that explicitly modeling boundary uncertainty is a powerful and promising direction for advancing temporal action localization.

\section{Related Work}
\subsection{Single-Stage Temporal Action Localization}
Early TAL methods often adopted a two-stage pipeline: first generating a sparse set of candidate temporal proposals, and then classifying and refining them \cite{b10, b11}. Models like BMN \cite{b10} achieved strong performance in proposal generation but were often computationally intensive and not end-to-end trainable. The field has since moved towards single-stage detectors, which perform localization and classification simultaneously in a single pass. A key innovation within this paradigm is the anchor-free approach, which discards pre-defined anchor windows and instead treats every temporal location on a feature map as a potential action candidate \cite{b2}. This design philosophy, exemplified by models like A2Net, GTAN, and notably ActionFormer, simplifies the architecture and has led to significant gains in both efficiency and accuracy \cite{b2}. TBT-Former builds directly upon this single-stage, anchor-free foundation, inheriting its efficiency while targeting its core representational limitations.

\subsection{Multi-Scale Representations in Video}
Actions in videos can span from a few seconds to several minutes, making multi-scale feature representation a critical component of any robust TAL model. The concept of a Feature Pyramid Network (FPN), originally developed for object detection, has been widely adopted to handle this variation \cite{b5}. An FPN creates a hierarchy of feature maps at different temporal resolutions, allowing the model to detect both short and long actions effectively. The baseline ActionFormer model implements a simple, feed-forward pyramid where deeper layers have coarser temporal resolution \cite{b2}. However, more advanced structures, such as the one explored in \cite{b5}, employ top-down pathways and lateral connections to explicitly merge features across scales. This allows high-level semantic information from coarse feature maps to flow back and enrich the spatially precise information in fine-grained maps, a principle we incorporate into our Cross-Scale FPN.

\subsection{Boundary Modeling in Detection}
A persistent challenge in localization tasks is the inherent ambiguity of object or action boundaries. Standard regression losses, such as L1 or Smooth L1, assume a single, deterministic ground-truth location, which is an oversimplification for many real-world cases. This issue was elegantly addressed in the object detection domain by Generalized Focal Loss (GFL) \cite{b6}. The core insight of GFL is to abandon the assumption of a Dirac delta distribution for a bounding box coordinate. Instead, it represents the coordinate as a flexible, arbitrary probability distribution learned over a set of discrete bins. The final coordinate is then computed as the expectation of this distribution, and a continuous version of Focal Loss is used for optimization \cite{b6}. This allows the model to represent its uncertainty; a sharp distribution indicates high confidence, while a flatter one signals ambiguity. In the TAL domain, TriDet \cite{b4} also explored modeling boundaries as distributions. Our proposed Boundary Distribution Regression (BDR) head is conceived as a direct and principled adaptation of the GFL philosophy to the temporal domain, providing a novel and effective solution for handling the pervasive issue of ambiguous action boundaries in videos.

\section{Overview of ActionFormer}
\begin{figure}[htbp]
\centering
\includegraphics[width=0.9\columnwidth]{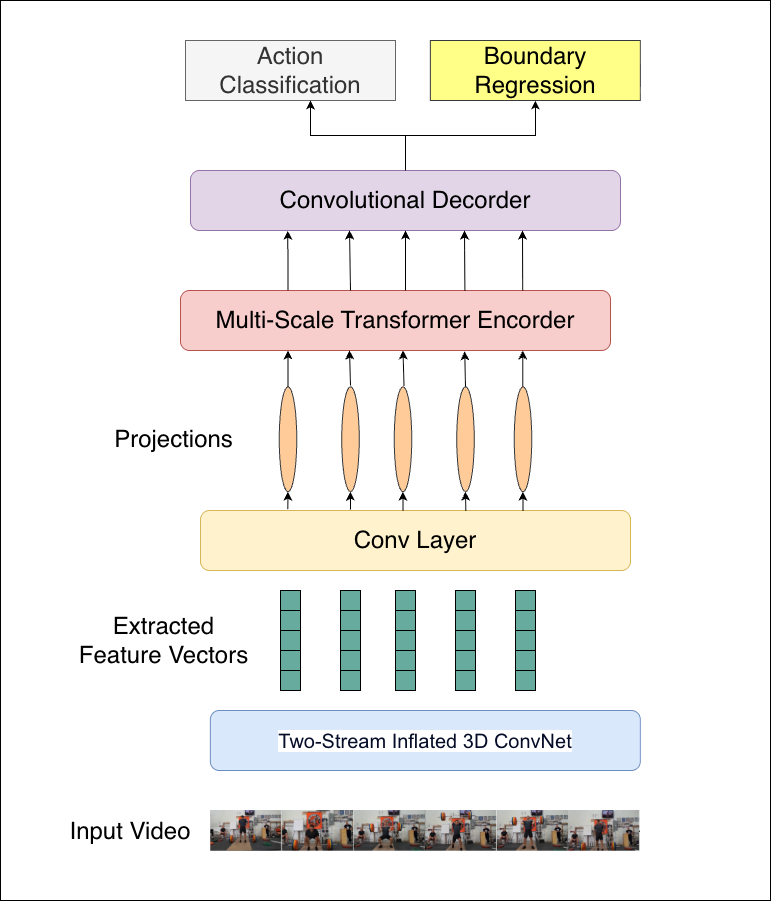}
\caption{An overview of the baseline ActionFormer architecture, illustrating the multi-scale Transformer encoder and the lightweight convolutional decoder with classification and regression heads \cite{b2}.}
\label{fig_actionformer}
\end{figure}

Our work builds upon the strong foundation of ActionFormer \cite{b2}, a minimalist and highly effective single-stage, anchor-free framework for temporal action localization. The core philosophy of ActionFormer is to treat every moment in a video as a potential action candidate, thereby converting the complex task of structured output prediction into a more manageable sequence labeling problem. An overview of the ActionFormer architecture is provided in Figure \ref{fig_actionformer}.

The model follows a standard encoder-decoder structure. The encoder is a multi-scale Transformer network that processes a sequence of input video features \cite{b2}. It utilizes local self-attention to model temporal dependencies efficiently, even for long videos. A key design choice is the creation of a feature pyramid, where strided operations within the Transformer blocks generate feature maps at progressively coarser temporal resolutions. This multi-scale representation allows the model to effectively detect actions of varying durations \cite{b2}.

The decoder consists of lightweight 1D convolutional networks that act as prediction heads. These heads are shared across all levels of the feature pyramid. For each temporal location on the pyramid, two parallel branches perform the main tasks: a classification head predicts the probability for each action class, and a regression head estimates the temporal distances from the current moment to the action's start and end boundaries. The model is trained end-to-end using a combination of Focal Loss \cite{b12} for classification and a DIoU loss \cite{b13} for boundary regression \cite{b2}. This simple yet powerful design established a new state-of-the-art, providing a robust baseline for further innovation in the field.

\section{Method}
\subsection{Architecture Overview}
\begin{figure}[htbp]
\centering
\includegraphics[width=0.9\columnwidth]{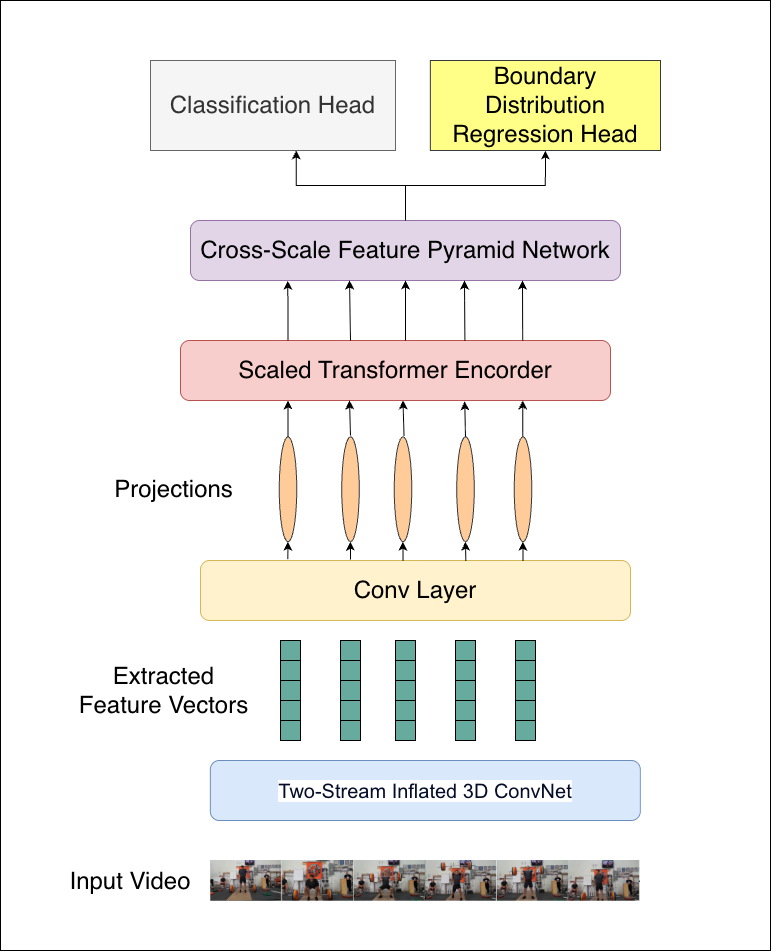}
\caption{Conceptual illustration of the TBT-Former architecture, highlighting the Scaled Backbone, Cross-Scale FPN (CS-FPN), and Boundary Distribution Regression (BDR) Head modules as key enhancements to the ActionFormer baseline.}
\label{fig_tbtformer}
\end{figure}

TBT-Former builds upon the robust and efficient anchor-free architecture of ActionFormer \cite{b2}. The baseline model consists of a Transformer-based encoder that processes a sequence of video features, followed by a lightweight convolutional decoder with separate classification and regression heads. Our proposed TBT-Former retains this overall structure but introduces three significant enhancements: a scaled-up backbone for increased capacity, a cross-scale feature pyramid for improved multi-scale fusion, and a novel regression head for probabilistic boundary modeling. Figure \ref{fig_tbtformer} illustrates the overall architecture of TBT-Former, highlighting our key modifications.

\subsection{Scaled Transformer Backbone}
\begin{figure}[htbp]
\centering
\includegraphics[width=0.8\columnwidth]{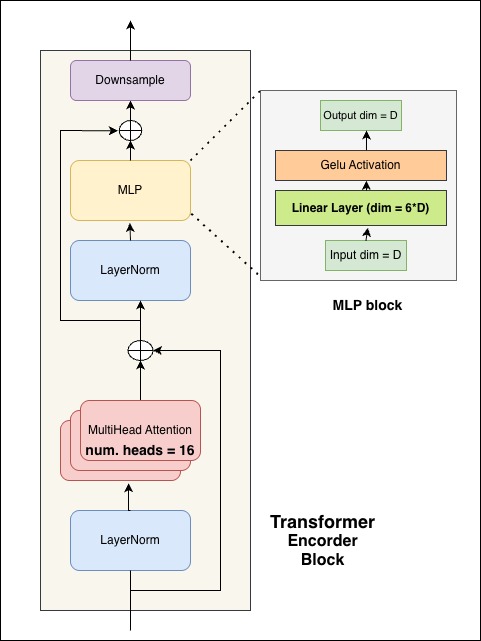}
\caption{The architecture of our Scaled Transformer Backbone unit. Compared to the baseline ActionFormer, we introduce two key enhancements for increased representational capacity: (1) the number of heads in the Multi-Head Attention module is increased to 16, and (2) the MLP block is expanded with a 6x hidden dimension, as shown in the zoomed-in view.}
\label{fig:scaled_backbone}
\end{figure}

To enhance the model's ability to learn complex temporal relationships and long-range dependencies, we first scale the capacity of the Transformer encoder. While the baseline ActionFormer uses a standard configuration for its Transformer blocks, we modify the key hyperparameters to create a more powerful backbone, as illustrated in Figure \ref{fig:scaled_backbone}. Specifically, within each Multi-Head Self-Attention (MSA) block, we increase the number of parallel attention heads from the baseline's 8 to 16. Furthermore, in the subsequent feed-forward network (FFN), we expand the hidden dimension by a factor of 6 relative to the input dimension, compared to the standard 4x expansion. This substantial increase in parameters provides the model with greater representational power, enabling it to capture more nuanced patterns in the temporal dynamics of actions, which is particularly beneficial for complex scenes and diverse datasets.

\subsection{Cross-Scale Feature Pyramid (CS-FPN)}
\begin{figure}[htbp]
\centering
\includegraphics[width=\columnwidth]{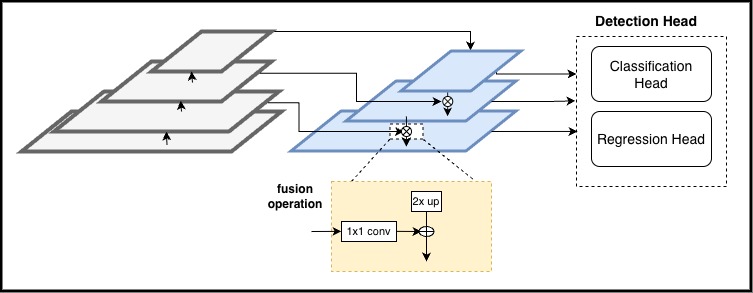}
\caption{The architecture of the Cross-Scale Feature Pyramid Network (CS-FPN). The fusion operation merges features from the bottom-up backbone pathway with the semantically richer features from the top-down pathway via lateral connections and element-wise addition.}
\label{fig:cs_fpn_fusion}
\end{figure}

The original ActionFormer builds its feature pyramid $\{P_2, P_3, P_4, P_5\}$ in a simple feed-forward manner, where each level is generated by downsampling the previous one \cite{b2}. This design prevents the flow of information from semantically rich, coarse-grained levels back to temporally precise, fine-grained levels. To remedy this, we introduce a Cross-Scale Feature Pyramid (CS-FPN), inspired by the design in \cite{b5}. The overall architecture and the specific fusion operation are illustrated in Figure \ref{fig:cs_fpn_fusion}.

Our CS-FPN enhances the feature hierarchy with a top-down pathway and lateral connections. Let $\{C_2, C_3, C_4, C_5\}$ be the output feature maps from different stages of our scaled backbone, with decreasing temporal resolution. The pyramid construction proceeds as follows:
\begin{enumerate}
    \item The coarsest feature map, $C_5$, is first passed through a $1 \times 1$ convolutional layer to produce the initial pyramid map, $P_5$.
    \item For each subsequent level $i$ (from 4 down to 2), the corresponding pyramid map $P_i$ is computed by merging the upsampled map from the coarser level, $P_{i+1}$, with the backbone map $C_i$. The merge operation is an element-wise addition:
    \begin{equation}
    P_i = \text{Conv}_{1 \times 1}(C_i) + \text{Upsample}(P_{i+1})
    \label{eq:fpn}
    \end{equation}
    where $\text{Conv}_{1 \times 1}$ is a lateral connection implemented as a 1D convolution to match channel dimensions, and $\text{Upsample}$ is a temporal upsampling operator (e.g., nearest-neighbor interpolation).
    \item Finally, each merged map $P_i$ is processed by a $3 \times 3$ convolution to generate the final set of feature maps for the prediction heads. This refinement step helps to reduce aliasing artifacts from the upsampling and merge operations.
\end{enumerate}
This architecture ensures that every level of the feature pyramid benefits from both high-resolution structural information (from the backbone) and high-level semantic context (from the top-down pathway), leading to a more powerful multi-scale representation for localizing actions of varying durations.

\subsection{Boundary Distribution Regression (BDR) Head}
Our most significant contribution is the Boundary Distribution Regression (BDR) head, which fundamentally changes how temporal boundaries are predicted.

\textbf{Motivation:} A standard regression head predicts a single float value for the start offset, $d^s$, and end offset, $d^e$. This formulation implicitly assumes that there is one precise, unambiguous boundary location. However, in reality, action boundaries are often fuzzy. Forcing the model to regress to a single point in such cases provides a noisy and unstable learning signal \cite{b4}.

\textbf{Probabilistic Formulation:} Inspired by Generalized Focal Loss (GFL) \cite{b6}, we reformulate this deterministic regression problem as a probabilistic one. Instead of predicting a single value $d^s$, our BDR head predicts a discrete probability distribution $P_s$ over a pre-defined range of $W$ possible integer offsets. The output for the start offset is a vector $P_s = \{p_s(0), p_s(1),..., p_s(W-1)\}$, where $\sum_{i=0}^{W-1} p_s(i) = 1$. The final continuous offset value, $\hat{d^s}$, is then calculated as the expectation of this distribution \cite{b6}:
\begin{equation}
\hat{d^s} = \sum_{i=0}^{W-1} i \cdot p_s(i)
\label{eq:bdr_exp}
\end{equation}
An identical process is used to predict the end offset distribution $P_e$ and compute $\hat{d^e}$. This formulation allows the model to express its uncertainty. A sharp, unimodal distribution for $P_s$ indicates high confidence in a specific boundary, while a flatter or multimodal distribution reflects ambiguity, which is a more faithful representation of the underlying data \cite{b6}.

\textbf{Loss Function:} To train the BDR head, we adapt the Distribution Focal Loss (DFL) from the GFL framework \cite{b6}. DFL provides an efficient and effective way to supervise the learning of this distribution. Given a continuous ground-truth offset $d^s_{gt}$, which lies between two discrete integer bins $i$ and $i+1$ (i.e., $i \le d^s_{gt} \le i+1$), DFL focuses the learning signal exclusively on these two adjacent bins. The loss is calculated as a weighted cross-entropy \cite{b6}:
\begin{equation}
\begin{split}
\mathcal{L}_{DFL}(P_s, d^s_{gt}) = -((i+1 - d^s_{gt})\log(p_s(i)) \\ + (d^s_{gt} - i)\log(p_s(i+1)))
\end{split}
\label{eq:dfl}
\end{equation}
This loss encourages the model to concentrate the probability mass of the distribution around the true continuous location. The total loss for our TBT-Former model is a combination of the standard Focal Loss ($\mathcal{L}_{cls}$) for classification and our DFL for the start and end boundaries:
\begin{equation}
\mathcal{L} = \mathcal{L}_{cls} + \lambda (\mathcal{L}_{DFL}(P_s, d^s_{gt}) + \mathcal{L}_{DFL}(P_e, d^e_{gt}))
\label{eq:total_loss}
\end{equation}
where $\lambda$ is a balancing hyperparameter. This loss structure provides a more stable and accurate learning objective for the challenging task of temporal boundary regression.

\section{Experiments}
\subsection{Datasets and Evaluation Metrics}
We evaluate TBT-Former on three widely-used and challenging TAL benchmarks:
\begin{itemize}
    \item \textbf{THUMOS14} \cite{b7}: This dataset is a standard benchmark for TAL, containing 20 action categories in over 400 untrimmed videos. It is characterized by long videos with a high density of action instances, making it particularly well-suited for evaluating localization precision.
    \item \textbf{ActivityNet-1.3} \cite{b9}: A large-scale dataset featuring 200 action classes in approximately 20,000 videos. Its scale and diversity make it an excellent testbed for model generalization and robustness.
    \item \textbf{EPIC-Kitchens 100} \cite{b8}: The largest dataset for egocentric action understanding. It contains numerous short, fine-grained actions defined by verb-noun pairs. Its unique characteristics test a model's ability to perform precise localization on brief, object-centric interactions.
\end{itemize}
For all datasets, we follow standard evaluation protocols and report the mean Average Precision (mAP) at various temporal Intersection over Union (tIoU) thresholds \cite{b2, b7}.

\subsection{Main Results}
We compare TBT-Former with the ActionFormer baseline and other state-of-the-art methods.

\textbf{Performance on THUMOS14:}
As shown in Table \ref{tab:thumos14}, TBT-Former establishes a new state-of-the-art on the THUMOS14 benchmark. It achieves an average mAP of 68.0\%, surpassing the strong ActionFormer baseline by 1.2 absolute points \cite{b2}. The inclusion of prior single-stage (GTAN, A2Net) and two-stage (BMN) methods further highlights the significant and consistent progress made, with TBT-Former outperforming all competitors across all tIoU thresholds \cite{b2}. This demonstrates the effectiveness of our architectural enhancements, particularly on a dataset that heavily rewards precise localization.

\begin{table*}[ht]
\caption{Comparison with state-of-the-art methods on the THUMOS14 test set. TBT-Former achieves the highest performance across all metrics.}
\label{tab:thumos14}
\centering
\begin{tabular}{@{}lccccccc@{}}
\toprule
Model & Type & Avg. mAP & mAP@0.3 & mAP@0.4 & mAP@0.5 & mAP@0.6 & mAP@0.7 \\
\midrule
GTAN \cite{b2} & Single-Stage & 57.8 & 70.4 & 63.3 & 57.8 & 47.2 & 33.5 \\
A2Net \cite{b2} & Single-Stage & 58.6 & 70.6 & 65.0 & 59.9 & 51.3 & 37.5 \\
BMN \cite{b2} & Two-Stage & 56.0 & 67.5 & 60.9 & 56.0 & 47.4 & 34.9 \\
ActionFormer \cite{b2} & Single-Stage & 66.8 & 82.1 & 77.8 & 71.0 & 59.4 & 43.9 \\
\textbf{TBT-Former (Ours)} & Single-Stage & \textbf{68.0} & \textbf{82.5} & \textbf{79.0} & \textbf{72.4} & \textbf{60.6} & \textbf{45.3} \\
\bottomrule
\end{tabular}
\end{table*}

\textbf{Performance on ActivityNet-1.3 and EPIC-Kitchens 100:}
Tables \ref{tab:activitynet} and \ref{tab:epickitchens} summarize our results on the ActivityNet-1.3 and EPIC-Kitchens 100 validation sets, respectively.

On the large-scale ActivityNet-1.3 dataset, TBT-Former achieves highly competitive performance. As shown in Table \ref{tab:activitynet}, our model is on par with the strong ActionFormer baseline, achieving an average mAP of 36.8\%. While there is a minor dip at the tIoU=0.5 threshold, our model shows improved performance at the stricter thresholds of tIoU=0.75 and tIoU=0.95 \cite{b2}. This indicates superior localization precision for well-defined actions and demonstrates that our architectural changes maintain strong performance on large-scale benchmarks.

On EPIC-Kitchens 100, presented in Table \ref{tab:epickitchens}, TBT-Former consistently outperforms ActionFormer and prior methods for both Verb and Noun classification, achieving average mAP improvements of 1.0\% and 1.2\% over ActionFormer, respectively \cite{b2}. The substantial gain, especially for Noun classification, suggests that our BDR head is particularly effective for localizing the short, object-centric actions that are prevalent in egocentric video.

\begin{table}[ht]
\caption{Performance on ActivityNet-1.3 validation set. The average mAP is calculated over [0.5:0.05:0.95]. *Results for ActionFormer are from the original paper \cite{b2}.}
\label{tab:activitynet}
\centering
\begin{tabular}{@{}lcccc@{}}
\toprule
Model & mAP@0.5 & mAP@0.75 & mAP@0.95 & Avg. mAP \\
\midrule
PGCN \cite{b3} & 48.3 & 33.2 & 3.3 & 31.1 \\
BMN \cite{b10} & 50.1 & 34.8 & 8.3 & 33.9 \\
G-TAD \cite{b11} & 50.4 & 34.6 & 9.0 & 34.1 \\
AFSD \cite{b15} & 52.4 & 35.3 & 6.5 & 34.4 \\
ActionFormer \cite{b2} & 54.7 & 37.8 & 8.4 & 36.6* \\
\textbf{TBT-Former (Ours)} & 53.9 & \textbf{38.2} & \textbf{8.5} & \textbf{36.8} \\
\bottomrule
\end{tabular}
\end{table}

\begin{table}[ht]
\caption{Performance on EPIC-Kitchens 100 validation set for Verb and Noun tasks. The average mAP is over [0.1:0.1:0.5].}
\label{tab:epickitchens}
\centering
\resizebox{\columnwidth}{!}{%
\begin{tabular}{@{}llcccccc@{}}
\toprule
Task & Model & mAP@0.1 & mAP@0.2 & mAP@0.3 & mAP@0.4 & mAP@0.5 & Avg. mAP \\
\midrule
\textbf{Verb} & BMN \cite{b10} & 10.8 & 9.8 & 8.4 & 7.1 & 5.6 & 8.4 \\
& G-TAD \cite{b11} & 12.1 & 11.0 & 9.4 & 8.1 & 6.5 & 9.4 \\
& ActionFormer \cite{b2} & 26.6 & 25.4 & 24.2 & 22.3 & 19.1 & 23.5 \\
& \textbf{TBT-Former (Ours)} & \textbf{27.2} & \textbf{26.7} & \textbf{25.6} & \textbf{22.9} & \textbf{20.1} & \textbf{24.5} \\
\midrule
\textbf{Noun} & BMN \cite{b10} & 10.3 & 8.3 & 6.2 & 4.5 & 3.4 & 6.5 \\
& G-TAD \cite{b11} & 11.0 & 10.0 & 8.6 & 7.0 & 5.4 & 8.4 \\
& ActionFormer \cite{b2} & 25.2 & 24.1 & 22.7 & 20.5 & 17.0 & 21.9 \\
& \textbf{TBT-Former (Ours)} & \textbf{26.4} & \textbf{25.9} & \textbf{23.8} & \textbf{21.7} & \textbf{17.9} & \textbf{23.1} \\
\bottomrule
\end{tabular}%
}
\end{table}

\subsection{Ablation Studies}
To validate our design choices and quantify the contribution of each new component, we conduct a systematic and extensive set of ablation studies on the THUMOS14 dataset.

\subsubsection{Summary of Component Contributions}
We first present a summary of how each of our proposed components contributes to the final performance. As shown in Table \ref{tab:ablation_summary}, each modification provides a clear and positive impact. The Scaled Backbone, CS-FPN, and BDR Head contribute +0.4, +0.3, and +0.8 mAP, respectively. When combined, they work synergistically to achieve a total improvement of +1.2 mAP over the strong ActionFormer baseline.

\begin{table}[ht]
\caption{Summary of component-wise contributions on THUMOS14.}
\label{tab:ablation_summary}
\centering
\begin{tabular}{@{}clcc@{}}
\toprule
\# & Model Configuration & Avg. mAP & $\Delta$ \\
\midrule
1 & Baseline (ActionFormer) & 66.8 & - \\
2 & + Scaled Backbone & 67.2 & +0.4 \\
3 & + Cross-Scale FPN (CS-FPN) & 67.1 & +0.3 \\
4 & + Boundary Distribution Head (BDR) & 67.6 & +0.8 \\
5 & \textbf{Full Model (TBT-Former)} & \textbf{68.0} & \textbf{+1.2} \\
\bottomrule
\end{tabular}
\end{table}

\subsubsection{Effect of Local Attention Window Size}
The original ActionFormer paper used a local attention window of size 19 for THUMOS14 \cite{b2}. Given that our Scaled Transformer Backbone has a higher representational capacity, we hypothesized that it could effectively leverage a larger temporal context. We experimented with several window sizes, with the results shown in Table \ref{tab:ablation_window}. Performance steadily improves as the window size increases from 19 to 30, where it peaks at an average mAP of 68.0\%. Beyond this, performance begins to plateau, suggesting that a window size of 30 provides the optimal balance between contextual reach and model focus for our enhanced architecture.

\begin{table}[ht]
\caption{Ablation on the local attention window size. All experiments use the full TBT-Former model.}
\label{tab:ablation_window}
\centering
\begin{tabular}{@{}lccc@{}}
\toprule
Window Size & mAP@0.5 & mAP@0.7 & Avg. mAP \\
\midrule
19 (Baseline size) & 71.5 & 44.2 & 67.1 \\
25 & 72.0 & 44.8 & 67.6 \\
\textbf{30} & \textbf{72.4} & \textbf{45.3} & \textbf{68.0} \\
37 & 72.2 & 45.1 & 67.8 \\
\bottomrule
\end{tabular}
\end{table}

\subsubsection{Design of the Feature Pyramid}
\begin{figure}[htbp]
\centering
\includegraphics[width=\columnwidth]{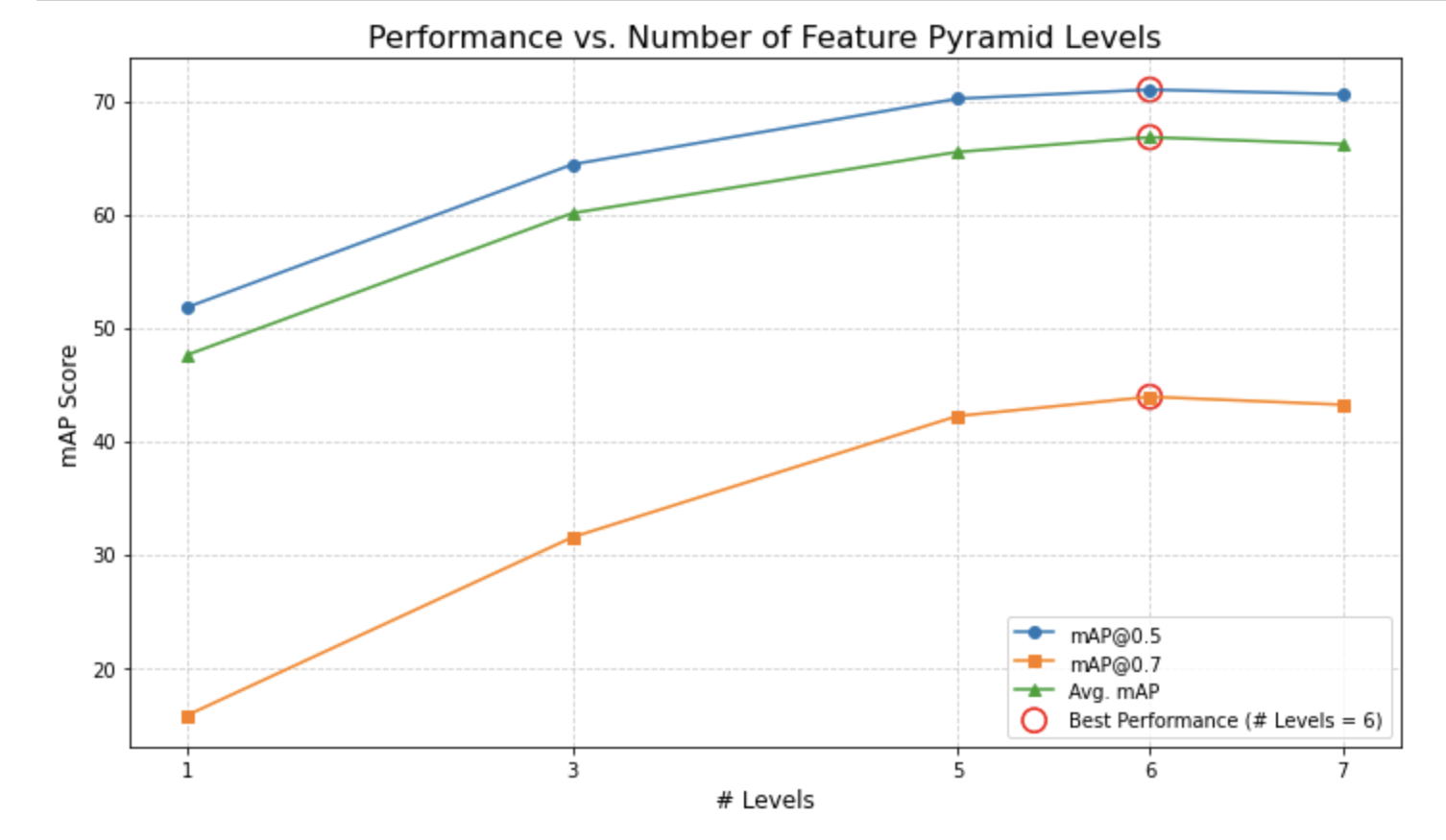}
\caption{Ablation on the number of feature pyramid levels. Performance across different metrics (mAP@0.5, mAP@0.7, and Avg. mAP) peaks at 6 levels, indicating this is the optimal depth for the pyramid in our architecture.}
\label{fig:ablation_fpn_plot}
\end{figure}

We validate the feature pyramid design inherited from ActionFormer \cite{b2}. As shown in Figure \ref{fig:ablation_fpn_plot}, performance is highly sensitive to the number of pyramid levels. A single-level model performs poorly (47.6\% avg. mAP). Performance consistently increases with more levels, saturating at 6 levels, which confirms that a deep, multi-scale pyramid is critical for capturing actions of varying durations.

\subsubsection{Analysis of Loss Weight \(\lambda_{reg}\)}
We analyze the balance between the classification loss and our new DFL regression loss, controlled by the hyperparameter $\lambda_{reg}$. As shown in Table \ref{tab:ablation_loss}, the model is robust to a range of values, but performance peaks at $\lambda_{reg}=1.0$. This confirms that an equal weighting between the classification and boundary distribution regression objectives provides the optimal trade-off for our model.

\begin{table}[ht]
\caption{Ablation on the regression loss weight $\lambda_{reg}$.}
\label{tab:ablation_loss}
\centering
\begin{tabular}{@{}lccc@{}}
\toprule
$\lambda_{reg}$ & mAP@0.5 & mAP@0.7 & Avg. mAP \\
\midrule
0.2 & 70.1 & 39.8 & 65.0 \\
0.5 & 71.4 & 41.7 & 66.4 \\
\textbf{1.0} & \textbf{71.0} & \textbf{43.6} & \textbf{66.9} \\
2.0 & 69.7 & 43.1 & 66.3 \\
5.0 & 68.8 & 42.5 & 65.1 \\
\bottomrule
\end{tabular}
\end{table}

\begin{figure*}[t]
    \centering
    \begin{subfigure}[b]{0.48\textwidth}
        \centering
        \includegraphics[width=\textwidth]{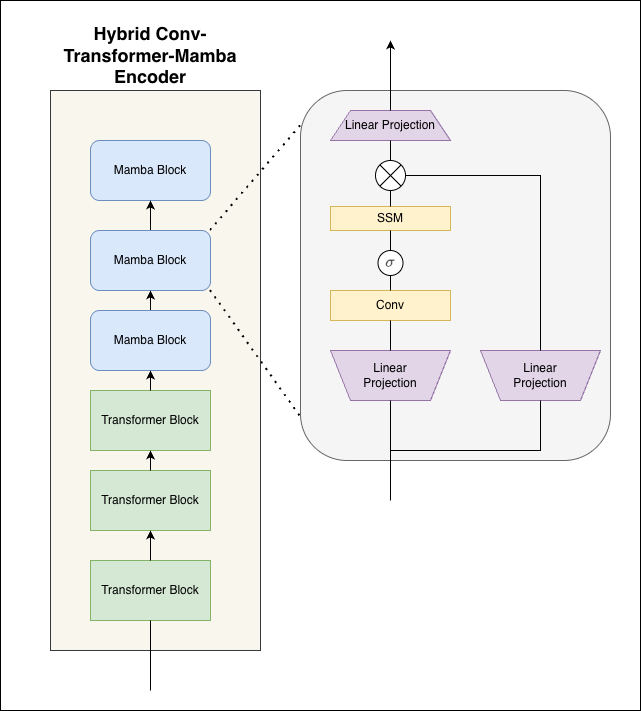}
        \caption{Hybrid Conv-Transformer-Mamba encoder.}
        \label{fig:hybrid_encoder_sub}
    \end{subfigure}
    \hfill
    \begin{subfigure}[b]{0.48\textwidth}
        \centering
        \includegraphics[width=\textwidth]{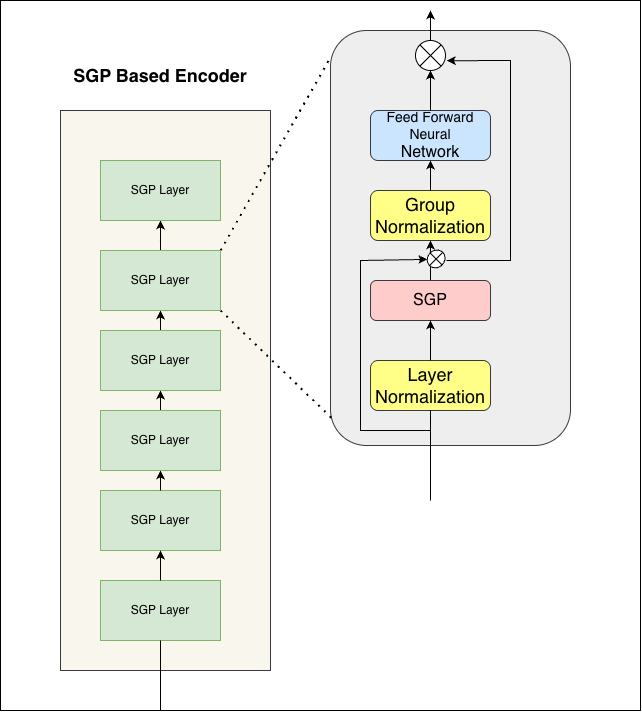}
        \caption{SGP-based encoder.}
        \label{fig:sgp_encoder_sub}
    \end{subfigure}
    \caption{Architectures of the alternative backbones explored in our ablation study. (a) The hybrid approach begins with convolutional embedding and an SE block, followed by a stem of Transformer blocks. The branch creates a feature pyramid using Transformer blocks at fine-grained levels and Mamba blocks at coarse-grained levels. (b) The SGP-based encoder is built from SGP blocks, with a two-branch design for feature discrimination and multi-granularity context.}
    \label{fig:alternative_backbones}
\end{figure*}

\subsubsection{Alternative Backbone Architectures}
Finally, we explored replacing the Transformer-based backbone of the baseline model with two alternative architectures to understand their suitability for TAL. The results are summarized in Table \ref{tab:ablation_backbones}.

\textbf{Hybrid Transformer-Mamba Backbone:} We designed a hybrid encoder that uses Transformer blocks for high-resolution feature maps and Mamba blocks for coarser, low-resolution maps. Mamba is a recent State Space Model (SSM) architecture with linear-time complexity, making it efficient for long sequences \cite{b14}. The core of Mamba is the Selective State Space Model (SSSM), which maps an input $x(t)$ to an output $y(t)$ via a latent state $h(t)$. Its discretized form is a recurrence:
\begin{equation}
\begin{split}
h_k = \bar{A}h_{k-1} + \bar{B}x_k \\
y_k = Ch_k
\end{split}
\label{eq:ssm_discrete}
\end{equation}
Mamba's innovation is making the parameters $\bar{B}$ and $\bar{C}$ input-dependent, allowing it to selectively focus on parts of the sequence \cite{b14}. Our hybrid architecture, shown in Figure \ref{fig:alternative_backbones}(a), achieved a competitive 65.8\% avg. mAP.

\textbf{SGP-based Backbone:} We also implemented a fully-convolutional backbone using the Scalable-Granularity Perception (SGP) layer from \cite{b4}. The SGP layer, shown in Figure \ref{fig:alternative_backbones}(b), replaces self-attention with two parallel branches: an instant-level branch to enhance feature discriminability and a window-level branch with parallel convolutions to capture multi-granularity context. The SGP operation is formulated as \cite{b4}:
\begin{equation}
f_{SGP}=\phi(x)FC(x)+\psi(x)(Conv_{w}(x)+Conv_{kw}(x))+x
\label{eq:sgp}
\end{equation}
where $\phi(x)$ and $\psi(x)$ are gating functions, and FC denotes a fully-connected layer. This model achieved 66.2\% avg. mAP.

While both alternatives show strong performance, they do not surpass the optimized Transformer baseline. This suggests that the local self-attention mechanism remains highly effective for temporal feature extraction in TAL, though the competitive results of SGP and Mamba highlight them as promising directions for future research.

\begin{table}[ht]
\caption{Ablation on alternative backbone architectures, compared to the ActionFormer baseline.}
\label{tab:ablation_backbones}
\centering
\begin{tabular}{@{}lccc@{}}
\toprule
Backbone Type & mAP@0.5 & mAP@0.7 & Avg. mAP \\
\midrule
Transformer (Baseline) & 71.0 & 43.9 & 66.8 \\
Hybrid Attn-Mamba & 69.5 & 42.9 & 65.8 \\
SGP (Convolutional) & 70.1 & 43.2 & 66.2 \\
\bottomrule
\end{tabular}
\end{table}

\section{Conclusion}
In this paper, we introduced TBT-Former, a novel single-stage, anchor-free model for temporal action localization. Our work makes three key contributions: a higher-capacity scaled Transformer backbone for enhanced temporal reasoning, a cross-scale feature pyramid for improved multi-scale fusion, and a novel Boundary Distribution Regression (BDR) head for modeling temporal ambiguity.

Within the paradigm of Transformer-based architectures, ActionFormer established a formidable performance benchmark. Our results demonstrate that TBT-Former successfully advances this benchmark, setting a new level of performance on the highly competitive THUMOS14 and EPIC-Kitchens 100 datasets, while remaining competitive on the large-scale ActivityNet-1.3. Our most significant finding is that explicitly modeling temporal boundaries as probability distributions is a powerful and promising paradigm.

Furthermore, our ablation studies on alternative backbones, such as the convolutional SGP and the SSM-based Mamba architectures, revealed their potential as efficient and competitive alternatives to self-attention. While they did not surpass our optimized Transformer-based model in this work, their strong performance indicates a valuable direction for future research. Subsequent work could focus on refining these architectures or developing more sophisticated hybrid models, potentially unlocking further gains in both performance and computational efficiency.

\end{document}